\documentclass[11pt,a4paper]{article}
\usepackage[hyperref]{naaclhlt2019}
\usepackage{times}
\usepackage{latexsym}
\usepackage{graphicx}
\usepackage{url}
\usepackage{booktabs}
\usepackage{amsmath}
\usepackage{soul}

\aclfinalcopy 
\def\aclpaperid{1760} 

\newcommand{\namedref}[2]{\hyperref[#2]{#1~\ref*{#2}}}

\title{Casting Light on Invisible Cities:\\ Computationally Engaging with Literary Criticism}

\author{Shufan Wang \\
  University of Massachusetts, Amherst \\
  {\tt shufanwang@cs.umass.edu} \\\And
  Mohit Iyyer \\
  University of Massachusetts, Amherst \\
  {\tt miyyer@cs.umass.edu} \\}

\date{}

\begin{document}
\maketitle
\begin{abstract}
Literary critics often attempt to uncover meaning in a single work of literature through careful reading and analysis. Applying natural language processing methods to aid in such literary analyses remains a challenge in digital humanities. While most previous work  focuses on ``distant reading'' by algorithmically discovering high-level patterns from large collections of literary works, here we sharpen the focus of our methods to a single literary theory about Italo Calvino's postmodern novel \textit{Invisible Cities}, which consists of 55 short descriptions of imaginary cities. Calvino has provided a classification of these cities into eleven thematic groups, but literary scholars disagree as to how trustworthy his categorization is. Due to the unique structure of this novel, we can computationally weigh in on this debate:  we leverage pretrained contextualized representations to embed each city's description and use unsupervised methods to cluster these embeddings. Additionally, we compare results of our computational approach to similarity judgments generated by human readers. Our work is a first step towards incorporating natural language processing into literary criticism.
\end{abstract}

\section{Introduction}

Literary critics form interpretations of meaning in works of literature. Building computational models that can help form and test these interpretations is a fundamental goal of digital humanities research~\cite{Benzen:76}. Within natural language processing, most previous work that engages with literature relies on ``distant reading''~\citep{Jockers:13}, which involves discovering high-level patterns from large collections of stories~\citep{Bamman, Chaturverdi}. We depart from this trend by showing that computational techniques can also engage with literary criticism at a closer distance: concretely, we use recent advances in text representation learning to test a single literary theory about the novel \textit{Invisible Cities} by Italo Calvino.

\begin{figure}[t!]\

\includegraphics[width=8.0cm]{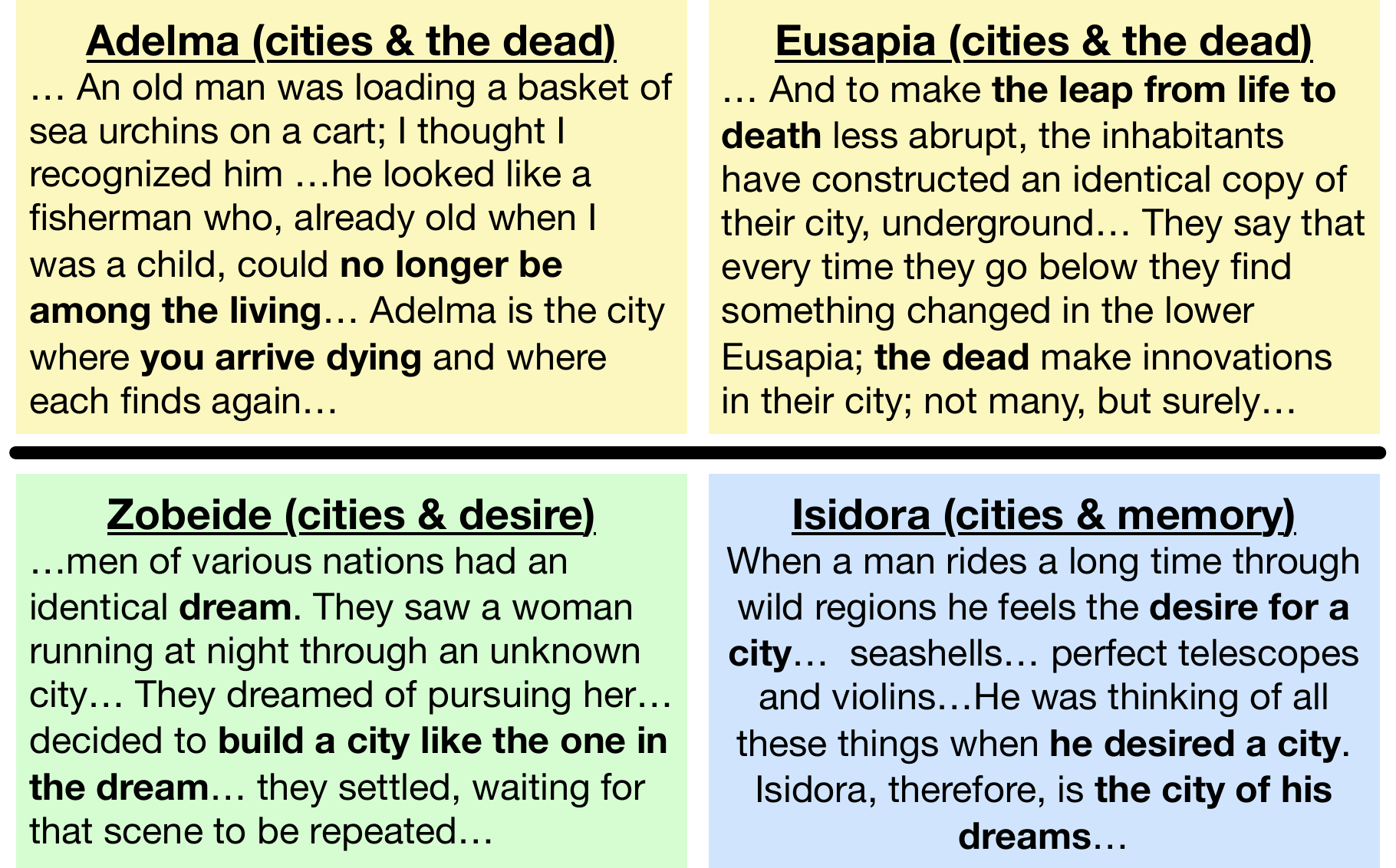}
\caption[descriptions]{Calvino labels the thematically-similar cities in the top row as \emph{cities \& the dead}. However, although the bottom two cities share a theme of desire, he assigns them to different groups.} \label{description}
\end{figure}

Framed as a dialogue between the traveler Marco Polo and the emperor Kublai Khan, \textit{Invisible Cities} consists of 55 prose poems,  each of which describes an imaginary city. Calvino categorizes these cities into eleven thematic groups that deal with human emotions (e.g., desires, memories), general objects (eyes, sky, signs), and unusual properties (continuous, hidden, thin). Many critics argue that Calvino's labels are not meaningful, while others believe that there is a distinct thematic separation between the groups, including the author himself~\cite{Calvino:04}. The unique structure of this novel --- each city's description is short and self-contained (Figure \ref{description}) --- allows us to computationally examine this debate.


As the book is too small to train any models, we leverage recent advances in large-scale language model-based representations~\cite{PetersELMo2018,devlin2018bert} to compute a representation of each city. We feed these representations into a clustering algorithm that produces exactly eleven clusters of five cities each and evaluate them against both Calvino's original labels and crowdsourced human judgments. While the overall correlation with Calvino's labels is low, both computers and humans can reliably identify some thematic groups associated with concrete objects.

While prior work has computationally analyzed a single book~\citep{Eve19}, our work goes beyond simple word frequency or n-gram counts by leveraging the power of pretrained language models to engage with literary criticism. Admittedly, our approach and evaluations are specific to \emph{Invisible Cities}, but we believe that similar analyses of more conventionally-structured novels could become possible as text representation methods improve. We also highlight two challenges of applying computational methods to literary criticisms: (1) text representation methods are imperfect, especially when given writing as complex as Calvino's; and (2) evaluation is difficult because there is no consensus among literary critics on a single ``correct'' interpretation. 



\section{Literary analyses of \emph{Invisible Cities}}

Before describing our method and results, we first review critical opinions on both sides of whether Calvino's thematic groups meaningfully characterize his city descriptions.

\paragraph{The groups are meaningful: } Some scholars believe that the thematic grouping imposed by Calvino reflects properties of the cities he describes; ~\citet{Dream}, for example, argues that Calvino's structure are ``ontologically grounded in different ways''. \citet{Port} further provides examples of cities with the same label that are clearly thematically similar, pointing at the ``cities of desire'' as ``informed by 20\textsuperscript{th} century theories of desires associated with Sigmund Freud''. \citet{Calvino:04} himself claims that he creates most categorizations of cities with clear labels in mind, especially the cities of memory and desire, which he deemed as ``fundamental cornerstones'' of the novel. However, many critics argue that authorial intent is irrelevant when analyzing literature~\citep{wimsatt1946intentional,barthes199411}.

\paragraph{The groups are arbitrary: }On the other hand, a large body of criticism focuses on the apparent mismatch between a city's assigned thematic group and the content of its descriptions. \citet{Bloom} claims that the ``cities are totally interchangeable''; \citet{Textual} agrees, stating that ``even the categories themselves seem both chosen and assigned arbitrarily''. \citet{UBC} contends that ``the catalogue is superimposed on, but does not cover, the elusive, fluid mass of an unwritten world''.

While out of scope for our computational analysis, many possible theories exist regarding \emph{why} the groupings appear largely incoherent. For instance, \citet{Tales} posits that the structural incoherence exists because all of the cities actually describe different facets of Marco Polo's hometown of Venice. \citet{emperor} argues instead that Calvino's labels ``may refer only to a projection of the Khan's occupational thirst for order, unrelated to the structure of the text'', while \citet{redemption} hypothesizes that the mismatch is one of many obstacles that readers need to  ``untangle'' to understand the central substance of the novel.

\section{A Computational Analysis}
\begin{figure}[t!]\
\begin{center}
    \includegraphics[width=\linewidth]{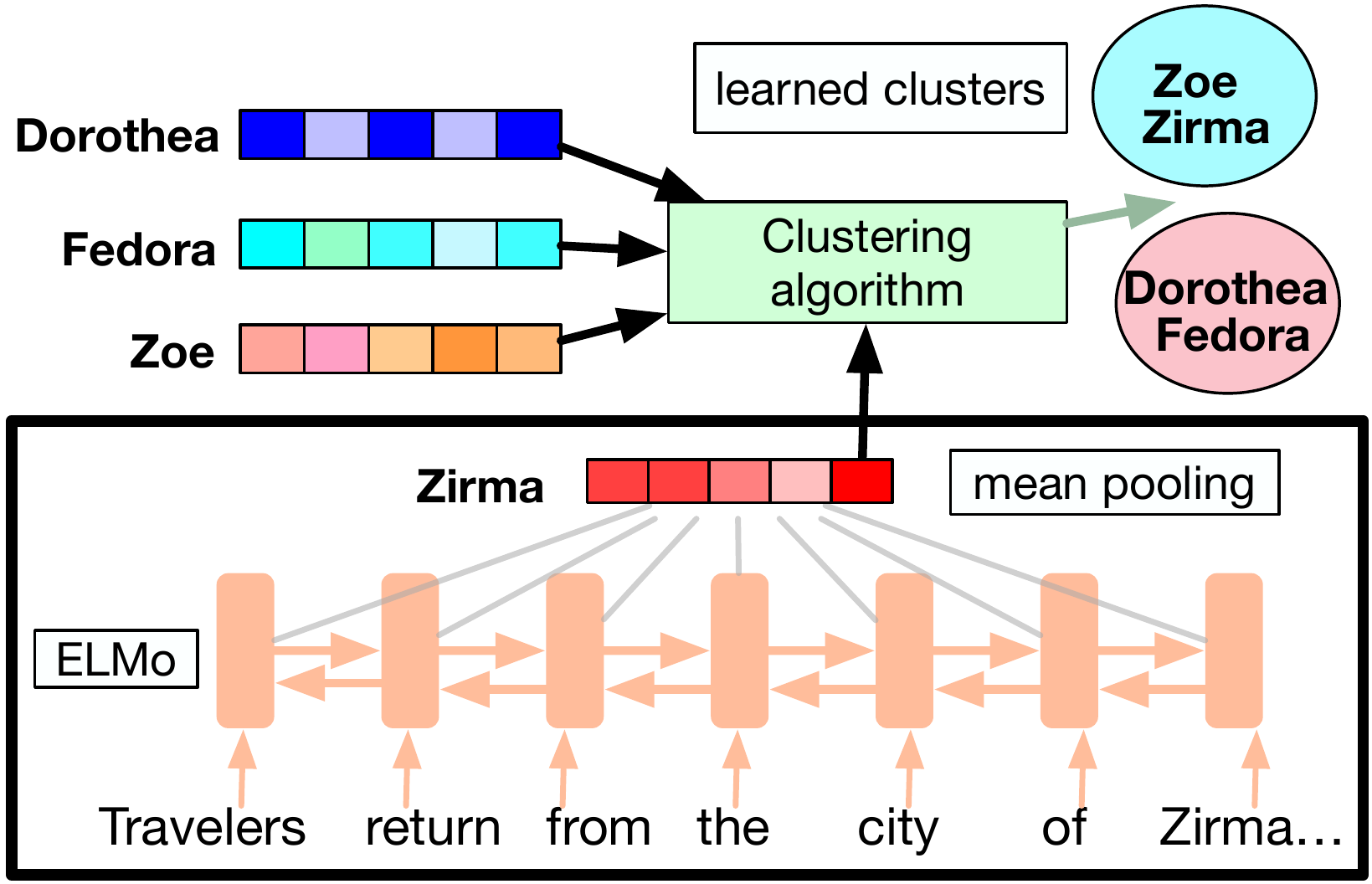}
    \caption{We first embed each city by averaging token representations derived from a pretrained model such as ELMo. Then, we feed the city embeddings to a clustering algorithm and analyze the learned clusters.} 
    \label{pipeline}
\end{center}
\end{figure}

We focus on measuring to what extent computers can recover Calvino's thematic groupings when given just raw text of the city descriptions. At a high level, our approach (Figure~\ref{pipeline}) involves (1) computing a vector representation for every city and (2) performing unsupervised clustering of these representations. The rest of this section describes both of these steps in more detail.

\subsection{Embedding city descriptions}
While each of the city descriptions is relatively short, Calvino's writing is filled with rare words, complex syntactic structures, and figurative language.\footnote{The book contains a vocabulary of 5,372 word types, and the average length of a city description is 380 tokens.}
Capturing the essential components of each city in a single vector is thus not as simple as it is with more standard forms of text. Nevertheless, we hope that representations from language models trained over billions of words of text can extract some meaningful semantics from these descriptions. We experiment with three different pretrained representations: ELMo~\citep{PetersELMo2018}, BERT~\citep{devlin2018bert}, and GloVe~\citep{pennington2014glove}. To produce a single city embedding, we compute the TF-IDF weighted element-wise mean of the token-level representations.\footnote{Using other composition functions such as the span representation of~\citet{Peters2018DissectingCW} had little impact on the learned clusters.} For all pretrained methods, we additionally reduce the dimensionality of the city embeddings to 40 using PCA for increased compatibility with our clustering algorithm. 



\subsection{Clustering city representations}
Given 55 city representations, how do we group them into eleven clusters of five cities each? 
Initially, we experimented with a graph-based community detection algorithm that maximizes cluster modularity \citep{PNAS}, but we found no simple way to constrain this method to produce a specific number of equally-sized clusters. 
The brute force approach of enumerating all possible cluster assignments is intractable given the large search space ($\frac{55!}{(5!)^{11}}$ possible assignments). We devise a simple clustering algorithm to approximate this process. First, we initialize with random cluster assignments and define ``cluster strength'' to be the relative difference between ``intra-group'' Euclidean distance and ``inter-group'' Euclidean distance.\footnote{The choice of distance metric (e.g., cosine, word mover) did not meaningfully impact our results.} Then, we iteratively propose random exchanges of memberships, only accepting these proposals when the cluster strength increases, until convergence. To evaluate the quality of the computationally-derived clusters against those of Calvino, we measure \emph{cluster purity}~\citep{Purity}:\footnote{Purity ranges between 0 and 1, and a larger purity indicates a higher degree of agreement.} given a set of predicted clusters $M$ and ground-truth clusters $D$ that both partition a set of $N$ data points, $$\text{purity} = \frac{1}{N}  \sum_{m \in M} max_{d \in D} | m \cap d |.$$


\begin{table}[t!]
\centering
\begin{tabular}{ccc}
  Method & Purity & Accuracy \\
  \toprule
   Random & 0.32 & 33.3\\
   GloVe & 0.35  & 35.9\\
   BERT & 0.40 & 39.3 \\
   ELMo & 0.42 & 44.6 \\
   Human & - & 48.8 \\
\end{tabular}
\caption{
 Results from cluster purity and accuracy on the ``odd-one-out'' task suggests that Calvino's thematic groups are not completely arbitrary.
  } \label{metrics}
\end{table}


\section{Evaluating clustering assignments}
While the results from the above section allow us to compare our three computational methods against each other, we additionally collect human judgments to further ground our results. In this section, we first describe our human experiment before quantitatively analyzing our results. 

\paragraph{Human clustering:}
We conduct a crowdsourced experiment to measure how well humans can disambiguate thematically different cities. Filling in the entire $55\times 55$ adjacency matrix with human similarity judgments is expensive and time-consuming. Thus, we instead design a proxy ``odd-one-out'' task for collecting human judgments: given three city descriptions, two of which come from the same ground-truth thematic group and the other from a different group, workers are asked to identify the \emph{intruder} city. We use the Figure Eight crowdsourcing platform\footnote{Workers were restricted to English-speaking countries and paid \$0.30 per judgment.} to collect three annotations each for 100 different city triples. 
Our interface initially displays only the first and last sentences of each city's description; workers can optionally click to reveal the full description. 
As workers are likely unfamiliar with \emph{Invisible Cities} and its different thematic groups, this crowdsourced task provides a fair comparison to our computational approaches.

\subsection{Quantitative comparison}
We compare clusters computed on different representations using community purity; additionally, we compare these computational methods to humans by their accuracy on the odd-one-out task. 

\paragraph{Purity of learned clusters: }
City representations computed using language model-based representation (ELMo and BERT) achieve significantly higher purity than a clustering induced from random representations, indicating that there is at least some meaningful coherence to Calvino's thematic groups (first row of Table~\ref{metrics}). ELMo representations yield the highest purity among the three methods, which is surprising as BERT is a bigger model trained on data from books (among other domains). Both ELMo and BERT outperform GloVe, which intuitively makes sense because the latter do not model the order or structure of the words in each description.

\paragraph{Comparison to humans: }
While the purity of our methods is higher than that of a random clustering, it is still far below 1. To provide additional context to these results, we now switch to our ``odd-one-out'' task and compare directly to human performance. For each triplet of cities, we identify the intruder as the city with the maximum Euclidean distance from the other two. Interestingly, crowd workers achieve only slightly higher accuracy than ELMo city representations; their interannotator agreement is also low,\footnote{Fleiss $\kappa=0.14$, indicating slight agreement, and two or more workers agreed on the intruder only 64\% of the time.} which indicates that close reading to analyze literary coherence between multiple texts is a difficult task, even for human annotators.
Overall, results from both computational and human approaches suggests that the author-assigned labels are not entirely arbitrary, as we can reliably recover some of the thematic groups. 


\section{Examining the learned clusters}
Our quantitative results suggest that while vector-based city representations capture some thematic similarities, there is much room for improvement. In this section, we first investigate whether the learned clusters provide evidence for any arguments put forth by literary critics on the novel. Then, we explore possible reasons that the learned clusters deviate from Calvino's.

\paragraph{Do learned clusters support existing analyses?}
The argument that \textit{cities of desire} constitute a particularly coherent thematic group~\citep{Port} is partially supported by our clustering results. Three of the five \textit{cities of desire} are grouped into the same cluster using BERT (two for ELMo), which makes it one of the most ``internally coherent'' groups. Similarly, some literary critics along with Calvino himself (\citealp{Calvino:04}) describe the \emph{thin cities} as a fairly arbitrary group, which is supported by our results: when using BERT, no two \emph{thin cities} are grouped into the same cluster. However, Calvino also suggests that the \textit{cities of memory} group is a ``fundamental substance'' of the book and therefore should be highly coherent. Our computational methods cannot pick up this theme, instead scattering all \textit{cities of memory} into different clusters.




\paragraph{Why do computers disagree with Calvino?}
In cases where the learned clusters deviate from the opinions of Calvino or literary critics, identifying the cause of the discrepancy is difficult: our computational methods are flawed, but there is also no one ``correct'' literary interpretation. Here we qualitatively analyze some of the learned clusters in an attempt to understand why the algorithm arrived at a particular assignment.
First, we examine two cities from different thematic groups, \emph{Beersheba} from ``cities and the sky'' and \emph{Valdrada} from ``cities and eyes'', that belong to the same learned cluster (and are each other's nearest neighbors).  The first two paragraphs of \emph{Beersheba} describe a noble city ``suspended in the heavens'' with an identical but immoral ``fecal'' city underground, while the remaining paragraphs focus on the heavenly city.  The description of \emph{Valdrada}, which is built on a lake, shares this theme of twin cities:  arriving travelers see ``two cities: one erect above the lake, and the other reflected, upside down''. While Calvino likely classified \emph{Beersheba} based on its location in the sky, the two cities share undeniable thematic similarities. Rerunning the clustering algorithm after removing the first two paragraphs of \emph{Beersheba} results in each city being assigned to a different cluster, which supports our hypothesis. 

Another interesting case is the previously-mentioned ``thin cities'', supposedly bound together by airy and ambiguous themes~\citep{redemption}, which \citet{Calvino:04} states were written after all of the other cities and are more incoherent than the other groups. While BERT does not group any thin cities together, ELMo categorizes \textit{Isaura} and \textit{Armilla} into the same learned cluster. The two cities appear largely dissimilar: \textit{Isaura} is a city with a thousand wells dug by its inhabitants, while \textit{Armilla} is an ``unfinished'' city without walls, ceilings, or floors. However,  both cities' descriptions mention supernatural beings living underground. In \textit{Isaura}, some people believe ``gods live in the depths'' and ``in the black lake that feeds the underground streams'', while the last paragraph of \textit{Armilla}'s description conjectures that it is ``in the possession of nymphs and naiads'' who ``travel along underground veins''. Removing these descriptions on underground gods and nymphs and rerunning our clustering algorithm yields a new assignment in which  each of these cities belongs to different clusters.

\paragraph{When do humans and computers agree?}
Our computational approach yields generally comparable accuracies and more consistent results than human annotators in the ``odd-one-out'' task. On cities with concrete themes such as sky and trading, our approach with BERT and ELMo obtains accuracy of 0.44 and 0.45 respectively, (0.47 and 0.48 for humans). ELMo also performs on par with humans in some case: for example, humans achieve an accuracy of 42\% on ``cities and eyes'', compared to ELMo's 43\%. On groups where the theme word frequently occurs in the passage, such as ``eyes'', our approach even slightly outperforms the human readers. However, human readers are better at recognizing abstract intangible topics, such as memory.



\section{Related work}
Most previous work within the NLP community applies distant reading \cite{Jockers:13} to large collections of books, focusing on modeling different aspects of narratives such as plots and event sequences~\citep{Jurafsky09,McIntyre,Goyal,Eisenberg17}, characters \citep{Bamman, Iyyer16,Chaturverdi16,Chaturverdi17}, and narrative similarity~\citep{Chaturverdi}.  In the same vein, researchers in computational literary analysis have combined statistical techniques and linguistics theories to perform quantitative analysis on large narrative texts~\citep{Michel11, Franzosi10, underwood16, Jockers16, Long16}, but these attempts largely rely on techniques such as word counting, topic modeling, and naive Bayes classifiers and are therefore not able to capture the meaning of sentences or paragraphs~\citep{Nanda}.  While these works discover general patterns from multiple literary works, we are the first to use cutting-edge NLP techniques to engage with specific literary criticism about a single narrative. 

There has been other computational work that focuses on just a single book or a small number of books, much of it focused on network analysis: ~\citet{Agarwal} extract character social networks from \emph{Alice in Wonderland}, while~\citet{ELSON} recover social networks from 19\textsuperscript{th} century British novels.~\citet{wallace2012multiple}  disentangles multiple narrative threads within the novel \emph{Infinite Jest}, while~\citet{Eve19} provides several automated statistical methods for close reading and test them on the award-winning novel \textit{Cloud Atlas} (2004). Compared to this work, we push further on modeling the content of the narrative by leveraging pretrained language models.

\section{Conclusion}
Our work takes a first step towards computationally engaging with literary criticism on a single book using state-of-the-art text representation methods. While we demonstrate that NLP techniques can be used to support literary analyses and obtain new insights, they also have clear limitations (e.g., in understanding abstract themes). As text representation methods become more powerful, we hope that (1) computational tools will become useful for analyzing novels with more conventional structures, and (2) literary criticism will be used as a testbed for evaluating representations.

\section*{Acknowledgement}
We thank the anonymous reviewers for their insightful comments. Additionally, we thank Nader Akoury, Garrett Bernstein, Chenghao Lv, Ari Kobren, Kalpesh Krishna, Saumya Lal, Tu Vu, Zhichao Yang, Mengxue Zhang and the UMass NLP group for suggestions that improved the paper's clarity, coverage of related work, and analysis experiments.











\bibliography{naaclhlt2019}
\bibliographystyle{acl_natbib}



\end{document}